\documentclass{article}

\usepackage{arxiv}

\usepackage[utf8]{inputenc}
\usepackage[T1]{fontenc}
\usepackage{lmodern}
\usepackage{amsmath}
\usepackage{amssymb}
\usepackage{geometry}
\usepackage{graphicx}
\usepackage{booktabs}
\usepackage{microtype}
\usepackage{hyperref}

\usepackage[square, numbers, sort&compress]{natbib}

\usepackage{listings}
\usepackage{xcolor}
\newcommand{\grad}{\nabla}
\newcommand{\R}{\mathbb{R}}
\newcommand{\Vocab}{\mathcal V}

\newtheorem{remark}{Remark}

\title{How Token Influence Decays with Distance:\\
A Green-Function View of Trained Language Models}
\author{
\href{https://orcid.org/0009-0009-4552-7424}{\includegraphics[scale=0.06]{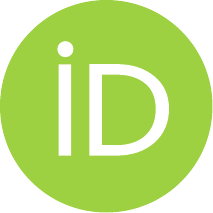}\hspace*{1mm}Matthias Brändel},
\href{https://orcid.org/0000-0003-1015-8736}{\includegraphics[scale=0.06]{orcid.pdf}\hspace*{1mm}{Stephan Köhler}}
and
\href{https://orcid.org/0000-0002-9310-8533}{\includegraphics[scale=0.06]{orcid.pdf}\hspace*{1mm}{Oliver Rheinbach}}\\
Faculty of Mathematics and Computer Science\\
Technische Universität Bergakademie Freiberg,\\ Freiberg, Germany}
\date{\today}

\begin{document}

\maketitle

%%The starting point of this work is a question from operator learning and scientific computing.

\begin{abstract}
We study how the next-token prediction of an autoregressive Transformer language model changes under small perturbations of earlier input token embeddings. Motivated by operator learning and iterative solvers for differential equations, we investigate how the influence of one token on another decays with distance in a trained model.
In multilevel methods for differential equations—such as domain decomposition, multigrid, and multilevel preconditioning—one often exploits a separation between strong local interactions and weaker but essential global interactions. The latter correspond to the long tail of the Green’s function and are typically handled by a coarse-level operator.
Inspired by this perspective, we compute an empirical, distance-resolved gradient profile of token dependencies using autograd. For a target position $j$ and distance $r$, we differentiate a selected next-token logit at position $j$ with respect to the input embedding at position $j-r$. This yields a local linearization of the learned nonlinear operator. The resulting object is not a Green’s function of the text itself, but a Green-function-like property of the trained model on a given context.
Experiments on trained Pythia models and Qwen2.5-0.5B show that, over the measured distance range, the median Jacobian sensitivity is much better described by a power-law-type decay than by an exponential alternative:
%Experiments on trained Pythia models and Qwen2.5-0.5B reveal that the median Jacobian sensitivity decays approximately as a power law with token distance:
the diagonal-normalized profile is well described by
\[
    \overline G(r) \approx \gamma+\beta(r+1)^{-p}
\]
with exponents $p \approx 0.7$–$0.9$ (typically $0.8$–$0.9$).

This behavior appears on coherent text from Gutenberg and WikiText-103. Token-shuffling experiments show that the power-law profile persists even when syntax and prediction quality collapse, whereas randomly initialized models do not exhibit it. The slowly decaying long-range sensitivity thus appears to be a learned property of trained autoregressive Transformer operators.
These findings suggest that hierarchical or coarse-level mechanisms in language models may be able to exploit the long-tailed sensitivity profiles.

\end{abstract}

\section{Introduction}

When striving to transfer ideas from %operator learning and
numerical analysis and scientific computing to natural language processing (NLP), a natural question emerges: how does the influence of one token on another decay with distance?  Large language models (LLMs) process sequences of tokens, and understanding this long-range information propagation is essential for interpretability, better architectural choices, and more
efficient training or inference.

The motivation for this question comes from hierarchical solvers for differential equations.
In a typical elliptic problem, local degrees of freedom interact strongly with nearby degrees of freedom, while global interactions are weaker but cannot be ignored.
In domain decomposition preconditioners, local subdomain solvers are therefore not enough: one also needs coarse functions that carry the global, slowly varying, or long-range part of the operator.
This principle appears already in classical Schwarz methods~\cite{Lions1988} and is central to modern domain decomposition and multigrid theory~\cite{ToselliWidlund2005,SmithBjorstadGropp1996,trottbook}.
In short, good hierarchical solvers exploit {strong local interactions} and represent {weak global interactions} by a suitable coarse operator.
Recent work at the interface of machine learning and domain decomposition further supports the idea that hierarchical numerical concepts can inform architectures in machine learning~\cite{Klawonn2024,Koehler:2026:HAD}.

Our recent operator-learning approach using linear attention was precisely of this type~\cite{Koehler:2026:HAD}.
If an attention-like architecture is interpreted as an approximation of an inverse operator, then a natural question is what the analog of coarse-grid functions should be for natural language.
For a Poisson problem, one has a clear mathematical answer: coarse basis functions, interface functions, or low-energy modes capture the global component.
For natural language processing, there is no obvious analog of a finite-element coarse basis.
Should the coarse variables correspond to words, sentences, topics, positions, summaries, or something else?  Before proposing such a construction, one should first ask whether trained language models even possess a measurable long-range response that would justify a coarse level.

This report attempts to address this question.  We use automatic differentiation to measure how selected next-token logits change when earlier input-token embeddings are infinitesimally perturbed.  Treating the forward pass of a trained LLM as a highly nonlinear operator, we empirically extract its local influence function across token distances.  This gradient-based analysis provides a quantitative and interpretable view of token interactions that goes
beyond conventional attention visualizations.  Attention weights describe one internal mechanism of the model, but they should not be identified directly with input-output explanations~\cite{jain2019attention,wiegreffe2019attention,abnar2020attentionflow}.  The derivative measured here describes the first-order effect of an input perturbation on an output logit.

%Our main empirical observation is that trained Transformer language models show slowly decaying long-range Jacobian sensitivities. This is in contrast to the typically exponential decay in classical recurrent neural networks (RNNs).

Our main empirical observation is that trained Transformer language models exhibit slowly decaying, long-range Jacobian sensitivities. This is in contrast to the typically exponential decay observed in classical recurrent neural networks (RNNs). %, which is consistent with their known limitations in capturing long-distance dependencies due to vanishing gradients
%
%
%It was our original goal to find a Green's function for text. However, our experiments showed that this is not quite what we have obtained.
%Token shuffling destroys coherent word order and makes next-token prediction much worse, but the %median distance
%token influence profile remains simular to the case of standard text.
%%approximately algebraic.
%We also find that randomly initialized Transformer models do
%not show the same distance profile.  Thus the measured object is not directly a property of natural language text, and it is also not a trivial property of the Transformer architecture. It is rather a property of the Transformer operator acquired during training on natural language.
%In this sense, we observe aspects of natural language through the lens of the trained transformer model.
%The text provides the point, where the model is evaluated,
%
%while training determines the local Jacobian response that we measure.
%
The decay of long-range token influence is stable across several corpora and models,
the median diagonal-normalized profile is well described, over the measured distance range, by a power-law-type decay rather than by the exponential alternative tested here.
%
%the median diagonal-normalized profile is well described by %algebraic
%a power law
%decay.
The effect persists when tokens are shuffled, but disappears for random weights.  This suggests that trained LLMs contain learned long-range response modes.
From the point of view of hierarchical architectures, these modes are precisely the kind of object that a coarse level would have to represent.

\section{From Green Functions to Transformer Jacobians}
In the theory of differential equations, for a linear problem \(Lu=f\), the Green's function~\cite{BookEvans2010} is the kernel of the
solution operator \(L^{-1}\):
$
u(x) = (L^{-1}f)(x) = \int {\mathcal G}(x,y) f(y)\,dy .
$
Formally, this means that
\[
\frac{\delta u(x)}{\delta f(y)} = {\mathcal G}(x,y),
\]
or, in a discrete problem \(Au=f\),
\[
\frac{\partial u_i}{\partial f_j} = (A^{-1})_{ij}.
\]
The columns of $A^{-1}$ are the discrete harmonic extensions (or equivalently, the energy-minimizing extensions) of unit perturbations on the right-hand side.

A Green's function therefore measures how strongly the solution at one position
responds to a localized perturbation of the input at another position. This sensitivity interpretation motivates our empirical analogue for language
models: we compute distance-resolved Jacobian responses of next-token logits with respect to earlier input embeddings.

We believe that the Green-function analogy may be useful in the present context because efficient hierarchical solvers for differential equations are designed to exploit the fact that Green's functions often exhibit strong local interactions together with weaker, but still important, long-range tails.

For an autoregressive Transformer, no fixed linear inverse matrix is available. We therefore transfer the response-based viewpoint to local Jacobian responses of the trained model.
A tokenized sequence with length $n$ is written as
\begin{equation}
T = (t_1,\ldots,t_n), \qquad t_i\in\Vocab,
\end{equation}
where $\Vocab$ is the vocabulary.
Let
\begin{equation}
  E=(e_1,\ldots,e_n), \qquad e_i\in\R^d,
\end{equation}
be the input embeddings of $T$, and let
\begin{equation}
  z_j(E)\in\R^{|\Vocab|}
\end{equation}
be the output logit vector at position $j$.
The vector $z_j(E)$ contains the scores used to predict the next token:
\begin{equation}
    \hat t_{j+1} = \arg\max_{v\in\Vocab} z_{j,v}(E).
\end{equation}

To obtain a scalar derivative, we select one component of this logit vector.
We use either the component corresponding to the observed next token in the sequence $t_{j+1}$ or the component corresponding to the model prediction:
\begin{align*}
  \phi_j^{\mathrm{true}}(E) &= z_{j,t_{j+1}}(E), \\
    \phi_j^{\mathrm{pred}}(E) &= z_{j,\hat t_{j+1}}(E).
\end{align*}

The distance-resolved sensitivity is defined for a target position $j$ and a causal token distance $r$ by
\begin{equation}
    G^{\alpha}(j,r)
  =
  \left\|
    \grad_{e_{j-r}} \phi_j(E)
  \right\|_2,
    \qquad
    \alpha\in\{\mathrm{true},\mathrm{pred}\}
\end{equation}
where $j-r$ is the source position. The parameters $j$ and $r$ are chosen such that $j-r$ lies inside the sequence.
The case $r=0$ measures the local self-sensitivity at the target position.
% Since the model is autoregressive, only source positions $j-r\leq j$ are considered. % already covered by r >=0
The Euclidean norm gives the
largest first-order change of the selected logit under a unit perturbation of
the source embedding.

% mb: gehört wenn dann irgendwo anders hin
% This is the precise sense in which we use the Green-function analogy.  We do
% not claim to compute a Green's function of language itself.  We compute a local
% Jacobian response of a trained nonlinear sequence operator.  In the linear case,
% that response would be a Green kernel entry.  In the Transformer case, it is an
% empirical Green-function diagnostic of the trained model.

The absolute scale of $G^{\alpha}(j,r)$ depends on embedding norms, logit scales, model
size, and the selected target position. Since the present study focuses on the
distance dependence of the response, each profile is normalized by its local
response at zero distance:
\begin{equation}
    \widetilde G^{\alpha}(j,r)
  =
  \frac{G^{\alpha}(j,r)}{G^{\alpha}(j,0)}.
\end{equation}
In the tables and figures, the corresponding normalized gradient-norm profile
based on the true next-token logit is denoted by \texttt{gradnorm\_diag};
the analogous profile based on the model-predicted next-token logit is denoted
by \texttt{pred\_gradnorm\_diag}.

%In the tables and figures, this quantity is denoted { gradnorm\_diag}.

Thus, for all retained normalized profiles, the local reference value satisfies
\begin{equation}
 \widetilde G^{\alpha}(j,0)=1.
\end{equation}

Let $\mathcal S$ denote the set of sampled sequences. For a fixed causal
distance $r$, the aggregate profile is defined by the median over all sampled
sequences $S$ and all target positions $j$:
\begin{equation}
    \overline G^{\alpha}(r)
  =
  \operatorname{median}_{S\in\mathcal S,\; j}
    \widetilde G^{\alpha}_S(j,r).
\end{equation}
The median represents the typical normalized response at distance $r$ and is
used because individual profiles can contain large outliers.
In the numerical implementation, profiles with non-finite $G^{\alpha}(j,0)$ or $|G^{\alpha}(j,0)|\leq\tau$, with
$\tau=10^{-30}$, are not included in the normalized aggregate statistics.

% We also computed logit-scale-normalized variants, in which the gradient norm is
% multiplied by the norm of the source embedding and divided by the standard
% deviation of the full logit vector at the target position. These variants led to
% the same qualitative conclusions. To keep the report focused, the main text
% concentrates on \texttt{gradnorm\_diag} and
% \texttt{pred\_gradnorm\_diag}.

All derivatives are computed by reverse-mode automatic differentiation in PyTorch~\cite{pytorch} with respect to the continuous input embeddings; see section~\ref{sec:autodiff}.

To characterize the distance dependence of the aggregated profile, we compare
three elementary decay models. We denote a fitted model profile by $m(r)$ and
use
\begin{equation}
  m_{\mathrm{pow}}(r)
  =
  \beta(r+1)^{-p},
\end{equation}
for a pure power law decay,
\begin{equation}
  m_{\mathrm{pow},\gamma}(r)
  =
  \gamma+\beta(r+1)^{-p},
\end{equation}
for a power law decay with a nonzero offset,
%background level,
and
\begin{equation}
  m_{\mathrm{exp},\gamma}(r)
  =
  \gamma+\beta\exp(-r/\xi),
\end{equation}
for an exponential decay with a characteristic length scale $\xi$.

The exponent $p$ measures the strength of the power law decay. Small values of
$p$ correspond to slowly decaying long-range influence. The parameter $\xi$
plays the analogous role for the exponential model: it is the distance scale over
which the response decreases by a factor of $e$. The offset $\gamma$ allows for a
distance-independent background level in the aggregated profile.

\begin{remark}
Since a Transformer defines a nonlinear map from the input embeddings to the next-token logits, there is no fixed Green's function in the classical linear sense. However, for each fixed input text, the model can be linearized around the corresponding embedding sequence. The Jacobian of a selected logit with respect to an earlier input embedding is then the analogue of the Green kernel of this local linearization. In our experiments we do not use the signed Jacobian entries themselves, but their embedding-space norms, aggregated as a function of the causal distance.
\end{remark}

\section{Related Work}
% Abschnitt zu: attention as explanation / attention is not explanation, gradient-based attribution in Transformers, influence functions oder Jacobian-/sensitivity-based analysis, long-range dependency in LMs, mechanistic interpretability, long-context/sparse/hierarchical Transformers.
%
% mechanistic interpretability
%
Our work is related to efforts in the %mechanistic %interpretability
analysis of Transformers. In particular, \cite{elhage2021framework} introduced the notion of an induction head, where two attention heads collaborate, e.g., for sequence completion, using context information over long distances.
In \cite{olsson2022incontext}, the authors adopt the idea of in-context learning as decreasing loss with increasing token index, which they attribute to \cite{kaplan2020scaling}.
%
%Our work is related to circuit-based analyses of Transformers. Elhage et al.~\cite{elhage2021framework} develop a mathematical framework for Transformer circuits and analyze how attention heads and MLPs can be decomposed into interpretable computational components.
%Within this line of work, Olsson et al.~\cite{olsson2022incontext} identify induction heads as a mechanism for sequence completion and in-context learning, building on the view that in-context learning can be observed as decreasing loss with increasing token position~\cite{kaplan2020scaling}.
%
% More recent circuit-tracing and information-flow methods construct attribution graphs or routes through the model to expose computational pathways underlying specific predictions~\cite{ameisen2025circuittracing,ferrando2024informationflow}.

These approaches aim to identify internal mechanisms or prediction-specific computational graphs. In contrast, our approach does not reconstruct individual circuits. We measure a distance-resolved Jacobian profile of the trained Transformer operator, aggregated over many contexts, in order to study how input-output sensitivity decays with causal token distance.
%induction heads as a key mechanism for in-context learning.
These works highlight the importance of understanding information flow and interaction patterns across positions. In contrast, our approach is motivated from multilevel methods in the numerical analysis of differential equations. % and studies global, distance-dependent properties of the trained model via gradient profiles.

% Our work is related to efforts in the %mechanistic %interpretability
% analysis of Transformers. In particular, \cite{elhage2021framework} introduced the notion of an induction head, where two attention heads collaborate, e.g., for sequence completion, using context information over long distances.
%
% In \cite{olsson2022incontext}, the authors adopt the idea of in-context learning as decreasing loss with increasing token index, which they attribute to \cite{kaplan2020scaling}.
% %“lsson et al.~\cite{olsson2022incontext} adopt a conception of in-context learning as decreasing loss with increasing token index, which they attribute to Kaplan et al.~\cite{kaplan2020scaling}.“
% %a mathematical framework for transformer circuits.
% They then propose that induction heads are a main mechanism realizing in-context and few-shot learning in transformers~\cite{olsson2022incontext}.
Our work also has some relation to the broader discussion of whether attention weights can be interpreted as explanations because gradients are typical alternatives.
In~\cite{jain2019attention}, the authors argue that attention weights should not automatically be treated as faithful explanations of model predictions; they consider the correlation of attention weights with gradient-based %measures of
feature importance.
The authors of \cite{wiegreffe2019attention} argue against this view and discuss conditions under which attention may still be explanatory.
The authors of \cite{bastings-filippova-2020-elephant} argue in favor of saliency methods instead of attention, which include gradient-based methods for the explanation of model predictions. Saliency methods are well established in computer vision and image classification. Such methods are closely related to our approach. However, we are not interested in explaining model predictions for specific inputs but rather in average sensitivity as a function of token distance.

In~\cite{abnar2020attentionflow}, the authors propose attention rollout and attention flow as methods for quantifying how information is mixed across layers in Transformers.
These approaches analyze attention-based internal mixing patterns.
In contrast, our diagnostic is based on derivatives of output logits with respect to input embeddings.
It therefore measures a first-order input-output sensitivity rather than an attention weight or an attention-flow score.
This places our method closer to gradient-based attribution methods such as Integrated Gradients~\cite{sundararajan2017integrated}, but, again, with a different objective:
we compute many gradient norms and aggregate them by causal distance using the median.

%we aggregate gradient norms. %sensitivities. % by causal token distance instead of attributing a single prediction to individual input tokens.

Recent related work includes~\cite{liu2026jacobianscopes,herasimchyk2026structural}.
In~\cite{liu2026jacobianscopes}, the authors propose to use
gradient information for attribution, i.e., to quantify how input tokens influence a model’s prediction. For example, using auto-diff on LlaMA-3.2 1B, they investigate how perturbations to an input token
embedding affect a hidden state.
As a gradient-based method, this work is clearly very closely related to ours. However, while in~\cite{liu2026jacobianscopes}, the authors are interested in performing attribution for a given specific context,
we are interested in averaged distance profiles of token dependencies using evaluation of the model at many different contexts.
In~\cite{herasimchyk2026structural}, the authors investigate influence profiles in Transformers where attention concentrates on early and late tokens to explain the {Lost-in-the-Middle} effect~\cite{liu2023lostmiddle} as a result of primacy and recency biases; see~\cite{herasimchyk2026structural} and the references therein.

Long-context behavior and position bias provide another important point of comparison.
The discussion on long-range effects go back to
\cite{sun2021longrange}, where the authors study whether long-range language models actually use distant context and show that the benefit of long context is selective rather than uniform across all predictions.

In~\cite{liu2023lostmiddle} the Lost-in-the-Middle effect was discussed showing that language models can fail to use relevant information when it is placed in the middle of a long context, even when the same information is used more reliably near the beginning or the end.
Relatedly,~\cite{xiao2024streamingllm} identifies attention sinks, where early tokens receive large attention mass even when they are not semantically important.

In~\cite{wu2025positionbias} and~\cite{herasimchyk2026structural} the authors analyze the emergence of position bias in Transformers and connect influence profiles to primacy and recency effects.
These works show that distance and position effects are fundamental in long-context Transformers.
The long-context and position-bias literature shows that distant tokens are not used uniformly and that models may prefer information near the beginning or the end of a context~\cite{sun2021longrange,liu2023lostmiddle,xiao2024streamingllm,herasimchyk2026structural}.

This supports the view that a coarse-level mechanism for language models should not be chosen solely from human semantic units such as sentences, paragraphs, or topics. It should also account for the learned long-range response structure of the trained operator. In our setting, this structure is represented by the distance-resolved Jacobian profile rather than by attention weights or retrieval accuracy alone.

%Other important approaches to understand Language models include graph-based approaches such as~\cite{ameisen2025circuittracing,ferrando2024informationflow}. However...
%These graphs are attribution-based, however, the methods are not based on gradients.
Other important approaches to understanding language models include graph-based methods in mechanistic interpretability, such as attribution graphs and information-flow routes~\cite{ameisen2025circuittracing,ferrando2024informationflow}.
These methods pursue a different goal than us: they aim to identify the internal components, features, or computational paths responsible for particular model outputs. In contrast, we do not attempt to reconstruct a circuit or assign attribution to individual heads, layers, or features. We use derivatives only as a local linear response of the original model and aggregate the resulting embedding-gradient norms by causal distance. %The object of study is therefore not a prompt-specific explanation, but a distance-resolved sensitivity profile of trained autoregressive language models.

\section{Automatic Differentiation}
\label{sec:autodiff}

The distance-resolved sensitivities \( G^{\alpha}(j,r) \) are computed via reverse-mode
automatic differentiation using PyTorch~\cite{pytorch}. Conceptually, the computation for one target position
\( j \) and causal distance \( r \) proceeds as follows:

\begin{lstlisting}[
    language=Python,
    basicstyle=\ttfamily\small,
    frame=single,
    breaklines=true,
    caption={Conceptual procedure for computing the empirical sensitivity \( G^{\alpha}(j,r) \).},
    label={lst:autograd}
]
# Conceptual algorithm (simplified for clarity)
embeds = model.get_input_embeddings()(input_ids)
embeds = embeds.detach().requires_grad_(True)      # make input embeddings differentiable

hidden = backbone(inputs_embeds=embeds)             # transformer forward pass
logits = lm_head(hidden)                            # [batch, seq_len, vocab_size]

# Select the scalar logit we want to differentiate
target_logit = logits[0, j, next_token_id]          # z_{j, t_{j+1}}

# Compute gradient of this scalar w.r.t. all input embeddings
grad = torch.autograd.grad(target_logit, embeds)[0] # shape: [batch, seq_len, d_model]

# Extract gradient at source position i = j-r and take L2 norm
sensitivity = grad[0, j - r].norm()                 # || \nabla_{e_{j-r}} z_j ||_2
\end{lstlisting}

In the actual implementation several practical optimizations are applied:
\begin{itemize}
    \item Only the required logit entry is computed (instead of materializing the full vocabulary-sized logit tensor).
    \item Multiple target positions are processed from a single forward pass.
    \item Gradients are computed for both the true next token and the model-predicted token.
    \item The resulting sensitivity profiles are normalized by their value at \( r = 0 \) and aggregated using the median across thousands of samples.
\end{itemize}

This yields the measured quantity
\[
    G^{\alpha}(j,r) = \bigl\| \nabla_{e_{j-r}} \phi_j(E) \bigr\|_2
\]
%exactly
as defined in Section~2.

\section{Experimental Setup}

We evaluate distance-resolved Jacobian profiles for trained autoregressive language models and for control experiments. The %main
model-size sweep uses the Pythia family~\cite{pythia2023}: Pythia-14M, Pythia-70M, Pythia-160M, and Pythia-410M. In addition, we evaluate Qwen2.5-0.5B as a model from a different trained model family~\cite{qwen2025qwen25}. As an architecture-only control, we also include a randomly initialized Pythia-14M model with the same configuration but untrained weights.

The main data source is a Gutenberg-style corpus of long documents. To test whether the observed profiles are specific to this text domain, we also evaluate WikiText-103~\cite{merity2016pointer}. % and a C++ code corpus.
As an order-destruction control, we use a shuffled Gutenberg variant. This preserves the token source but destroys the coherent sequential structure of the text.

The datasets are not evaluated exhaustively.
Instead, a bounded set of candidate documents is first loaded.
At most 1500 candidate documents are retained.
The order of these candidate documents is randomized before tokenization.
Documents with fewer than $L+1$ tokens are discarded, where $L$ is the model input context length.
From the remaining tokenized documents, the scripts draw $1024$ contiguous token chunks with replacement.
Each chunk has length $L+1$.
The first $L$ tokens are used as model input, and the additional token provides the observed next-token target for the last admissible target position.
For the Pythia runs, the model input context length is $L=2048$. For Qwen2.5-0.5B, we use $L=4096$.

%The datasets are not evaluated exhaustively.
%Instead, first a bounded set of candidate documents is loaded.
%At most 1500 candidate documents are retained.
%This candidate set is then shuffled and tokenized.
%Documents with fewer than $L+1$ tokens are discarded, where $L$ is the model input context length.
%From the remaining tokenized documents, the scripts draw $1024$ contiguous token chunks with replacement.
%Each chunk has length $L+1$.
%The first $L$ tokens are used as model input, and the additional token provides the observed next-token target for the last admissible target position.
%For the Pythia runs, the model input context length is $L=2048$. For Qwen2.5-0.5B, we use $L=4096$.

% All target-position parameters in this section refer to zero-based implementation indices, with $j_{\max}$ excluded.
% Within each sampled input sequence, target positions are selected close to the
% right end of the context. For the Pythia runs, the target-position parameters
% are $j_{\min}=1920$, $j_{\max}=2048$, $j_{\Delta}=8$.
% For Qwen2.5-0.5B, the corresponding parameters are $j_{\min}=3584$, $j_{\max}=4096$, $j_{\Delta}=32$.

For each selected target position $j$, gradients are computed for prescribed
causal distances $r$. For the Pythia runs
as well as the Qwen2.5-0.5B run,
the distance values are
\begin{equation}
\begin{split}
  r\in\{&0,1,2,3,4,5,6,8,10,12,16,24,32,48,64,96,128,192,\\
       &256,384,512,768,1024,1280,1536\}.
\end{split}
\end{equation}
%Qwen2.5-0.5B uses the same distance values.
%For Qwen2.5-0.5B, the distance values are
%\begin{equation}
%\begin{split}
%  r\in\{&0,1,2,3,4,5,6,8,10,12,16,24,32,48,64,96,128,192,\\
%       &256,384,512,768,1024,1536,2048\}.
%\end{split}
%\end{equation} % vielleicht sagen wie viele rows das dann sind
% TODO depends on which run we use, mine has
% # oliver
% # PYTHIA_R_VALUES="0,1,2,3,4,5,6,8,10,12,16,24,32,48,64,96,128,192,256,384,512,768,1024,1280,1536"
% # QWEN_R_VALUES="0,1,2,3,4,5,6,8,10,12,16,24,32,48,64,96,128,192,256,384,512,768,1024,1536,2048"
% # matthias
% PYTHIA_R_VALUES="0,1,2,3,4,5,6,8,10,12,16,24,32,48,64,96,128,192,256,384,512,768,1024,1280,1536,1792"
% QWEN_R_VALUES="0,1,2,3,4,5,6,8,10,12,16,24,32,48,64,96,128,192,256,384,512,768,1024,1280,1536,1792,2048,2308,2560,2816,3072,3326"

All target-position parameters in this section refer to zero-based implementation
indices, with $j_{\max}$ excluded.  Within each sampled input sequence,
target positions are selected close to the right end of the context.  For the
Pythia runs, the target-position parameters are
$j_{\min}=1920$, $j_{\max}=2048$, and $j_{\Delta}=8$.  This gives
$16$ target positions per sampled chunk.

%Target positions are deliberately restricted to a narrow interval near the end of the context, and the maximum evaluated distance is chosen such that even for the largest $r$, the corresponding source position remains well inside the available context. This reduces potential confounding by absolute-position effects such as primacy and recency biases~\cite{sun2021longrange,liu2023lostmiddle,xiao2024streamingllm,herasimchyk2026structural} and allows the measured profiles to more closely reflect dependence on causal distance alone.

Target positions are deliberately restricted to the same narrow interval near
the end of each context window.  This keeps the absolute target positions and
the prediction setting comparable across all evaluated distances, avoids mixing
different target regions of the context, and the maximum evaluated distance is
chosen such that even for the largest $r$, the corresponding source position
remains well inside the available context.
This reduces potential confounding by absolute-position effects such as primacy
and recency biases~\cite{sun2021longrange,liu2023lostmiddle,xiao2024streamingllm,herasimchyk2026structural}
and allows the measured profiles to more closely reflect dependence on causal
distance alone.
However, source-position effects are reduced but not fully removed.

Since we use $1024$ sampled
chunks, each Pythia run contains
$$
1024 \times 16 = 16384
$$
evaluated sample--target-position pairs over which $\overline G^{\alpha}(r)$ is computed.  For Qwen2.5-0.5B, the corresponding
parameters are $j_{\min}=3584$, $j_{\max}=4096$, and $j_{\Delta}=32$,
again giving $16$ target positions per chunk and thus $16384$ evaluated
sample--target-position pairs.

The decay models introduced above are fitted to the aggregated median profiles.
Fits are performed in log-space.
The value $r=0$ is used for normalization but excluded from the decay fits.
Model comparison uses log-space  root mean squared error (RMSE) for the fit. % and AIC.
%The log-space RMSE measures the relative quality of the fit.
%, while the AIC additionally penalizes the number of free parameters.

\section{Results}

\subsection{Prediction quality at measured target positions}

This section is a sanity check for the selected target logits.
It shows whether the true-token logits used in the sensitivity computation correspond to plausible model predictions, and it separates normal language-modeling cases from diagnostic controls such as shuffled text and randomly initialized models.

Table~\ref{tab:prediction-quality-all} reports prediction quality at the target positions.
All entries are computed over $16384$ evaluated target positions per experiment.
For each target position, the model assigns a logit score to every vocabulary token as a possible next token.
Sorting these scores in descending order gives the rank of the actually observed next token.

Top-$k$ accuracy is the fraction of target positions for which the observed next token appears among the $k$ highest-scoring entries of the model's logit vector.

\begin{table}[thbp]
\centering
\resizebox{\textwidth}{!}{%
\begin{tabular}{lrrrrr}
\toprule
Experiment
& Top-1
& Top-5
& Top-10
& Median rank
& Mean rank \\
\midrule
Pythia-14M, Gutenberg
& 26.77\% & 45.61\% & 53.50\% & 5 & 297.5 \\
Pythia-70M, Gutenberg
& 33.15\% & 54.16\% & 62.00\% & 3 & 146.5 \\
Pythia-160M, Gutenberg
& 40.53\% & 62.97\% & 70.50\% & 2 & 111.9 \\
Pythia-410M, Gutenberg
& 45.70\% & 67.88\% & 75.51\% & 2 & 80.8 \\
Qwen2.5-0.5B, Gutenberg
& 42.54\% & 66.50\% & 74.05\% & 2 & 113.1 \\
\midrule
Pythia-14M, WikiText-103
& 25.06\% & 43.36\% & -- & 5 & 404.2 \\
% Pythia-14M, C++ code
% & 65.84\% & 80.49\% & 83.64\% & 1 & 91.0 \\
\midrule
Pythia-14M, shuffled Gutenberg
& 3.39\% & 9.69\% & -- & 234 & 3084.8 \\
\midrule
Random-init Pythia-14M, Gutenberg
& 0.00\% & 0.01\% & -- & 23287.5 & 23799.9 \\
Random-init Pythia-14M, shuffled Gutenberg
& 0.00\% & 0.00\% & -- & 23371.0 & 24122.4 \\
\bottomrule
\end{tabular}%
}
\caption{Prediction quality at the target positions. All entries are computed
over $16384$ evaluated target positions per experiment. A dash indicates that
the corresponding value was not recorded in that run.}
\label{tab:prediction-quality-all}
\end{table}

For coherent Gutenberg text, prediction quality improves with model size.
Top-1 accuracy increases from $26.77\%$ for Pythia-14M to $45.70\%$ for Pythia-410M, and the median true-token rank decreases from $5$ to $2$.
Qwen2.5-0.5B reaches a comparable prediction quality on the same corpus, with median rank $2$ and top-10 accuracy $74.05\%$.
% The C++ code corpus is substantially easier for Pythia-14M than prose.
%The model reaches $65.84\%$ top-1 accuracy and median rank $1$, which is plausible because code contains strong local regularities and frequent syntactic tokens.
%\resizebox{\textwidth}{!}{%
%\begin{tabular}{lrrrrrr}
%\toprule
%Experiment
%& Rows
%& Top-1
%& Top-5
%& Top-10
%& Median rank
%& Mean rank \\
%\midrule
%Pythia-14M, Gutenberg
%& 16384 & 26.77\% & 45.61\% & 53.50\% & 5 & 297.5 \\
%Pythia-70M, Gutenberg
%& 16384 & 33.15\% & 54.16\% & 62.00\% & 3 & 146.5 \\
%Pythia-160M, Gutenberg
%& 16384 & 40.53\% & 62.97\% & 70.50\% & 2 & 111.9 \\
%Pythia-410M, Gutenberg
%& 16384 & 45.70\% & 67.88\% & 75.51\% & 2 & 80.8 \\
%Qwen2.5-0.5B, Gutenberg
%& 16384 & 42.54\% & 66.50\% & 74.05\% & 2 & 113.1 \\
%Pythia-14M, WikiText-103
%& 16384 & 25.06\% & 43.36\% & -- & 5 & 404.2 \\
%Pythia-14M, C++ code
%& 16384 & 65.84\% & 80.49\% & 83.64\% & 1 & 91.0 \\
%Pythia-14M, shuffled Gutenberg
%& 16384 & 3.39\% & 9.69\% & -- & 234 & 3084.8 \\
%Random-init Pythia-14M, Gutenberg
%& 16384 & 0.00\% & 0.01\% & -- & 23287.5 & 23799.9 \\
%Random-init Pythia-14M, shuffled Gutenberg
%& 16384 & 0.00\% & 0.00\% & -- & 23371.0 & 24122.4 \\
%\bottomrule
%\end{tabular}%
%}
%\end{table}

The shuffled Gutenberg control is different.  After shuffling, word order,
syntax, and long-range semantic structure are destroyed.  The trained Pythia-14M
model reaches only $3.39\%$ top-1 accuracy and $9.69\%$ top-5 accuracy.  In this
control, even a unigram baseline, i.e., predicting tokens solely by their global frequency in the corpus, is stronger:
$
  \text{unigram Top-1}=4.24\%,
  %\qquad
  \text{unigram Top-5}=17.14\%,
  %\qquad
  \text{unigram Top-10}=26.79\%. % TODO: Warum geben wir das mit, wir vergleichem it Pythio-14M, shuffled Gutenberg und da sind auch nir top-1 und top-5 gegeben
$
Thus the shuffled experiment should not be interpreted as successful language
modeling.  It is used to test %a diagnostic tool for
the trained operator under a deliberately
non-linguistic input distribution.

%The shuffled Gutenberg control is different.
%After shuffling, word order, syntax, and long-range semantic structure are destroyed.
%The trained Pythia-14M model reaches only $3.39\%$ top-1 accuracy and $9.69\%$ top-5 accuracy, with median rank $234$.
%A unigram baseline ignores the context and ranks candidate next tokens only by their empirical frequency in the corpus.
%In the shuffled Gutenberg control, this context-free baseline is already stronger than the trained Pythia-14M model: $\text{unigram Top-1}=4.24\%,\ \text{Top-5}=17.14\%$.
%Thus the shuffled experiment should not be interpreted as successful language modeling.
%It is a diagnostic for the trained operator under a deliberately non-coherent input distribution.

The randomly initialized Pythia-14M model performs much worse still, with essentially zero accuracy.

These values are not intended as benchmark results for the model families.
They report prediction quality only at the target positions used for the Jacobian measurements.
Published model reports for Pythia and Qwen2.5 provide standard benchmark evaluations, but those evaluations are not directly comparable to the present target-position Top-$k$ statistics.

%**hier**
\begin{remark}
Token shuffling does not remove all information from the input.  The shuffled sequence still contains the same tokens. %, and therefore preserves lexical, topical, and distributional cues such as names, objects, and recurring expressions.
A %instruction-tuned
language model may use such cues to produce a plausible thematic summary, although much of the apparent narrative structure may then come from the model's prior rather than from the shuffled sequence itself.  Thus, the shuffled experiment is not a test against a semantically empty input.  Rather, it tests whether the distance-resolved Jacobian profile depends on coherent sequential word order and successful next-token prediction.
The collapse in prediction accuracy under shuffling shows that the autoregressive task is strongly disrupted. At the same time, the persistence of a similar sensitivity profile indicates that the measured long-range structure is not determined solely by coherent next-token prediction.
%The collapse in prediction accuracy under shuffling shows that the autoregressive task is strongly disrupted, while the persistence of a similar sensitivity profile indicates that the measured long-range structure is not simply a proxy for coherent next-token prediction.
\end{remark}

\subsection{Distance-Resolved Jacobian Profiles}

%Table~\ref{tab:gradnorm-diag-all-fits} summarizes the AIC-selected fits for the median \texttt{gradnorm\_diag} profiles.
%For the trained models, the AIC-selected fits favor a power law decay over the exponential alternative considered here.
%Depending on the run, the selected power law model is either a pure power law or a power law with a small empirical floor.
%algebraic
%decay over exponential decay.  In several cases the selected model is

%The selected model is the one with the smallest log-space RMSE; in the 14M case, pure power and power with floor are indistinguishable, and we report the simpler pure power model.
%****

Tables~\ref{tab:gradnorm-diag-all-fits} and~\ref{tab:gradnorm-diag-all-fits_quality} report the fitted decay models for the
% median \texttt{gradnorm\_diag}
median profiles $\overline G^{\alpha}(r)$, together with their log-space RMSE as well as the quality of the fits.
Here, the true next token is used.
Across the trained models, the exponential-with-floor model has a substantially
larger log-space RMSE than
the power law
alternatives.
Thus, over the measured
distance range, the median profiles are much better described by a power-law-type
decay than
%by an exponential decay.
by the exponential-with-floor model considered here.

For Pythia-70M, Pythia-160M, Pythia-410M, and Qwen2.5-0.5B, the best fit in
terms of log-space RMSE is a power law with a small empirical floor,
\[
    \overline G^{\alpha}(r) = \gamma + \beta (r+1)^{-p}
\]
For Pythia-14M, the pure power law and the power law with floor give the same
log-space RMSE up to the reported precision, and the fitted floor is
numerically zero.  In this case, we therefore report the simpler pure power-law
fit.  We emphasize that the floor is an empirical parameter over the finite
measured distance range, not evidence for a nonzero asymptotic limit.

%****
%Table~\ref{tab:gradnorm-diag-all-fits} summarizes the fits for the median
%\texttt{gradnorm\_diag} profiles, evaluated by the log-space RMSE.
%For all trained models, the power law models provide a substantially better
%description of the measured distance profiles than the exponential-with-floor
%alternative considered here.
%For Pythia-70M, Pythia-160M, Pythia-410M, and Qwen2.5-0.5B, the smallest
%log-space RMSE is obtained by a power law with a small empirical floor,
%\[
%  g(r) = A r^{-p} + c .
%\]
%For Pythia-14M, the pure power law and the power law with floor have
%indistinguishable log-space RMSE, and the fitted floor is numerically zero.
%We therefore report the simpler pure power-law model in this case.

\begin{table}[htbp]
\centering
\resizebox{\textwidth}{!}{%
\begin{tabular}{llrrrr}
\toprule
Experiment
& Best model
& $\beta$
& $p$
& $\gamma$
& logRMSE \\
\midrule
Pythia-14M, Gutenberg
& pure power
& 0.5877 & 0.8825 & -- & 0.1614 \\
Pythia-70M, Gutenberg
& power + floor
& 1.1049 & 0.8817 & 0.000971 & 0.0607 \\
Pythia-160M, Gutenberg
& power + floor
& 1.1700 & 0.8193 & 0.004573 & 0.0355 \\
Pythia-410M, Gutenberg
& power + floor
& 1.0120 & 0.7148 & 0.010076 & 0.0594 \\
Qwen2.5-0.5B, Gutenberg
& power + floor
& 1.1628 & 0.8811 & 0.001787 & 0.0690 \\
Pythia-14M, WikiText-103
& pure power
& 0.5767 & 0.8725 & -- & 0.1454 \\
% Pythia-14M, C++ code
% & pure power
% & 0.8095 & 0.9105 & -- & 0.1255 \\
Pythia-14M, shuffled Gutenberg
& pure power
& 0.5114 & 0.8372 & -- & 0.1382 \\
Random-init Pythia-14M, Gutenberg
& degenerate pure power
& 0.0007649 & $\approx 0$ & -- & 0.1256 \\
Random-init Pythia-14M, shuffled Gutenberg
& degenerate pure power
& 0.0007633 & $\approx 0$ & -- & 0.1264 \\
\bottomrule
\end{tabular}%
}
\caption{Best fits for the median \texttt{gradnorm\_diag} profiles.  For the
random-init controls, the fitted exponent is essentially zero; the formally
selected pure-power fit is therefore a degenerate constant profile.}
\label{tab:gradnorm-diag-all-fits}
\end{table}

% aus fits_14m*490m.txt
\begin{table}[t]
\centering
%\begin{tabular}{lcccc|c}
\begin{tabular}{lcccc}
\hline
Model & Best model by logRMSE
    & $m_{\mathrm{pow}}$ & $m_{\mathrm{pow},\gamma}$ & $m_{\mathrm{exp},\gamma}$
% & power law exponent $p$ \\ % TODO: do we need this last column?
\\
\hline
Pythia-14M
& pure power / power+floor
& 0.1614--0.1622
& 0.1614--0.1622
& 0.6220--0.6229
% & 0.8816--0.8825 \\
\\
Pythia-70M
& power+floor
& 0.0906--0.0932
& 0.0607--0.0618
& 0.4861--0.4866
% & 0.8817--0.8843 \\
\\
Pythia-160M
& power+floor
& 0.1335--0.1355
& 0.0340--0.0355
& 0.3527--0.3549
% & 0.8193--0.8212 \\
\\
Pythia-410M
& power+floor
& 0.1343--0.1383
& 0.0594--0.0615
& 0.2940--0.2955
% & 0.7134--0.7148 \\
\\
Qwen2.5-0.5B
& power+floor
& 0.1349--0.1380
& 0.0682--0.0690
& 0.4198--0.4227
% & 0.8811--0.8869 \\
\\
\hline
\end{tabular}
    \caption{Fit quality of the median distance profiles $\overline G^{\mathrm{true}}(r)$ and $\overline G^{\mathrm{pred}}(r)$.
  The entries give ranges of log-space RMSE over
    these two quantities.  The $m_{\mathrm{exp},\gamma}$ model is consistently worse.
    Except for Pythia-14M, where the fitted floor in the $m_{\mathrm{pow},\gamma}$
model is of order \(10^{-12}\) and hence the fit effectively reduces to a
pure power law, the smallest log-space RMSE is obtained by a power law with
an empirical floor.
%Except for Pythia-14M, where the fitted floor in the power-law-with-floor model is numerically negligible, the smallest log-space RMSE is obtained by a power law with an empirical floor.
}
\label{tab:gradnorm-diag-all-fits_quality}
\end{table}

For Pythia-14M, the fitted exponent is stable across different input types:
\[
  p_{\mathrm{Gutenberg}}\approx 0.88,
  \qquad
  p_{\mathrm{WikiText}}\approx 0.87,
  % \qquad
  % p_{\mathrm{C++}}\approx 0.91.
\]
This indicates that the effect is not specific to one corpus.
% The C++ result is particularly useful because the prediction problem is much easier there, yet the measured decay exponent is close to the prose experiments.

Figure~\ref{fig:median_true} shows the median Jacobian sensitivity for Qwen2.5-0.5B in a log-log-plot.
Figure \ref{fig:individual_true} depicts randomly selected 250 individual profiles as well as the median, mean and trim10; for each causal distance, trim10 is computed after discarding the lowest and highest 10\% of the finite positive gradient-norm values and averaging the remaining values.

Figure~\ref{fig:pythia_median_true} shows the same quantities as in Figure~\ref{fig:median_true}, but for the largest Pythia model with 410M parameters.

\IfFileExists{figures/qwen490m_gradnorm_diag_aggregate_fit_loglog.png}{%
\begin{figure}[htbp]
\centering
\includegraphics[width=0.9\textwidth]{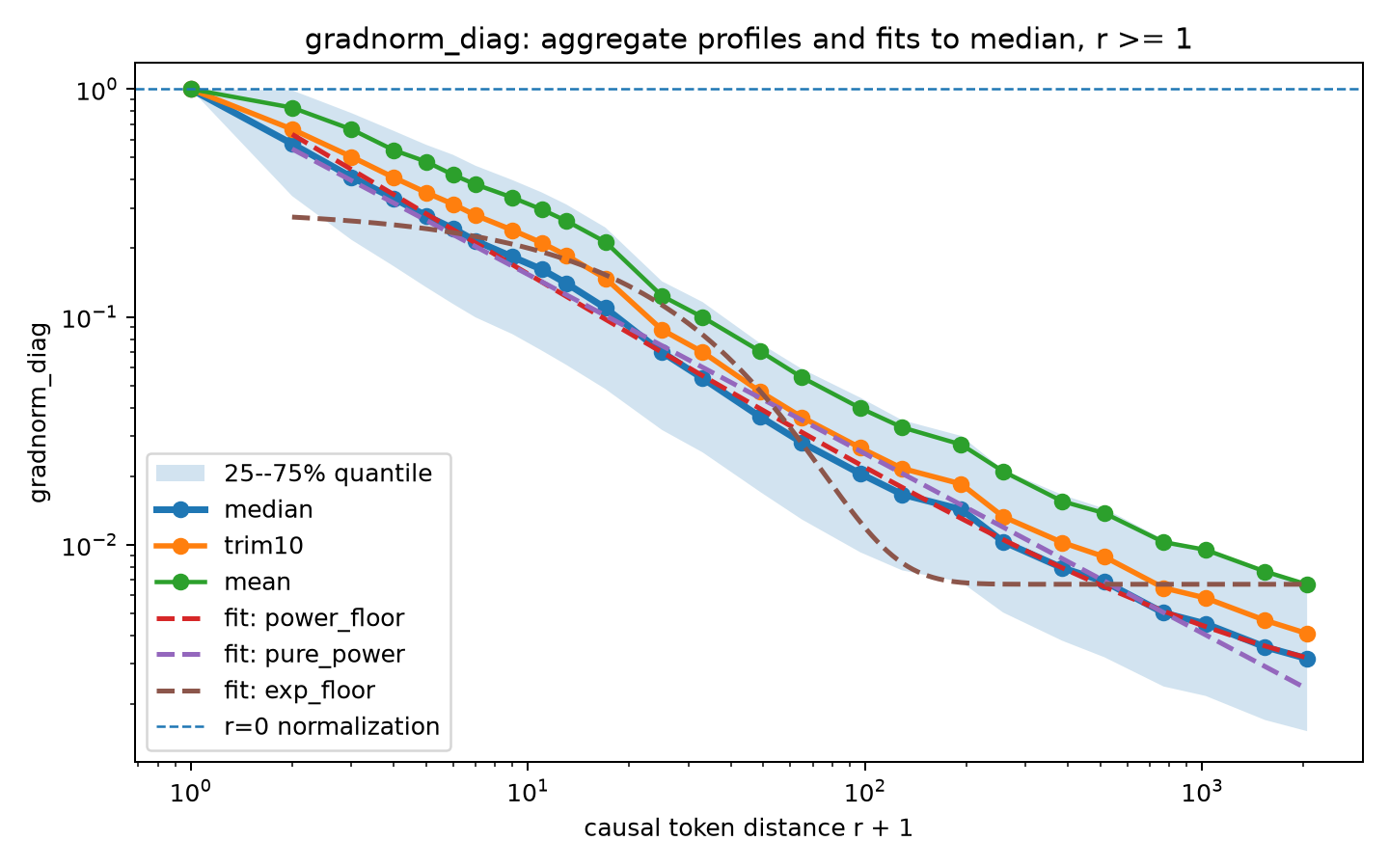}
\caption{Median distance-resolved Jacobian sensitivity profile for Qwen2.5-0.5B using the true next-token logit.}
\label{fig:median_true}
\end{figure}
}{}

\IfFileExists{figures/qwen490m_gradnorm_diag_individual_loglog.png}{%
\begin{figure}[htbp]
\centering
\includegraphics[width=0.9\textwidth]{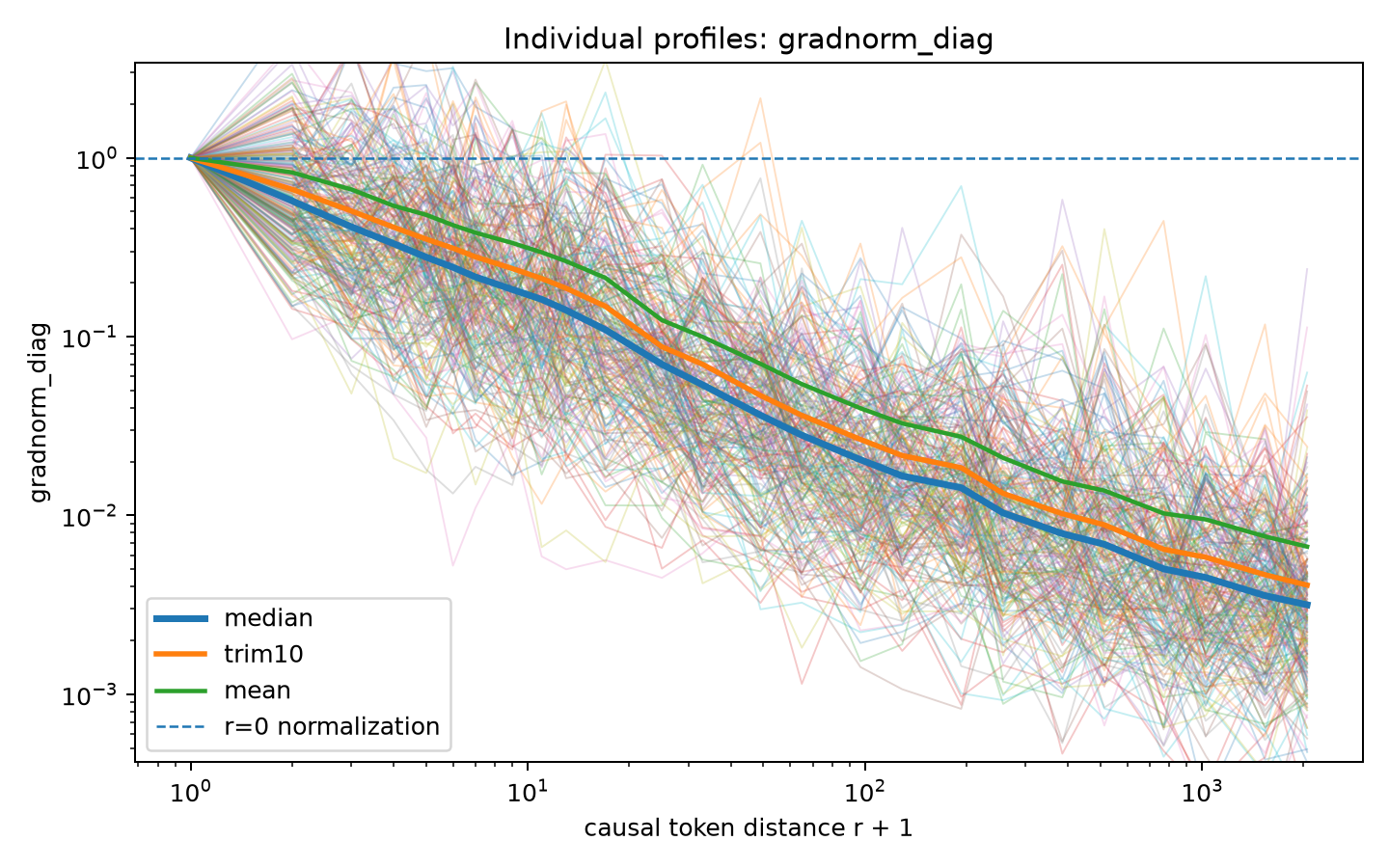}
\caption{Individual diagonal-normalized sensitivity profiles for Qwen2.5-0.5B using the true next-token logit.  The spread illustrates why medians are used as the primary aggregation statistic.}
\label{fig:individual_true}
\end{figure}
}{}

\IfFileExists{figures/pythia410m_gradnorm_diag_aggregate_fit_loglog.png}{%
\begin{figure}[htbp]
\centering
\includegraphics[width=0.9\textwidth]{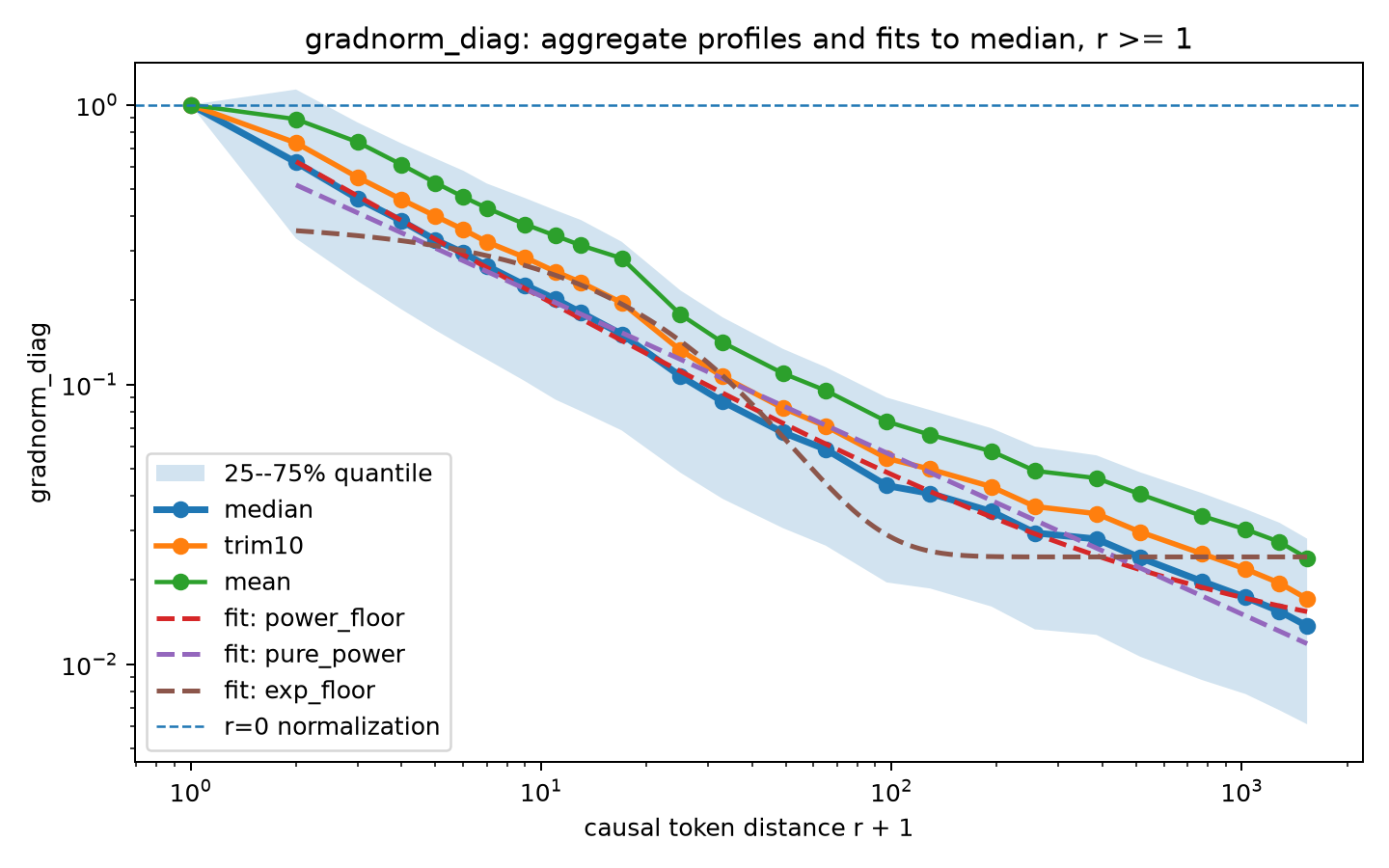}
\caption{Median distance-resolved Jacobian sensitivity profile for Pythia-410M using the true next-token logit.}
\label{fig:pythia_median_true}
\end{figure}
}{}

\subsection{True-token versus predicted-token logits}

\begin{figure}[ht]
\centering
\includegraphics[width=0.9\textwidth]{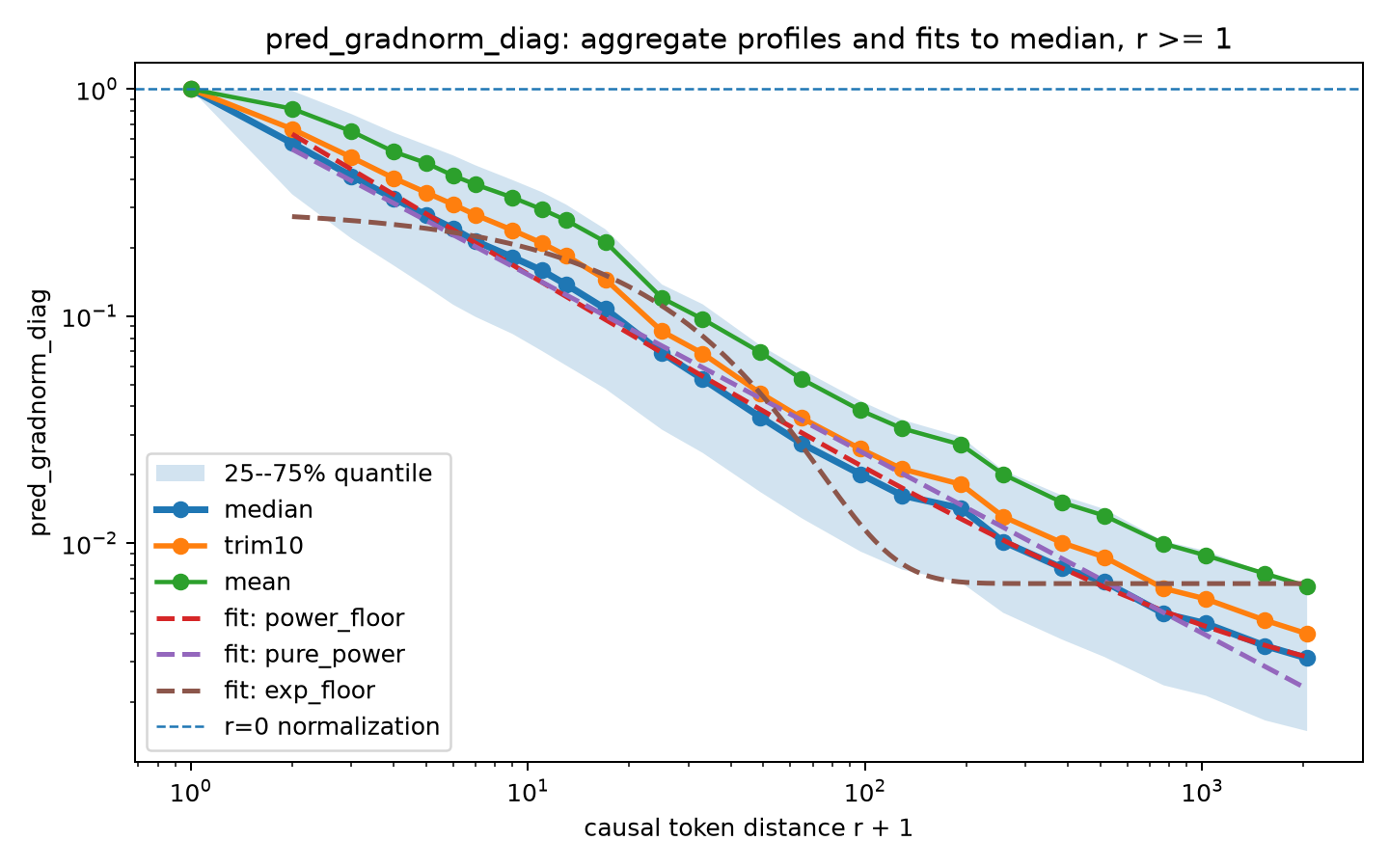}
\caption{Median distance-resolved sensitivity profile for Qwen2.5-0.5B using the model-predicted next-token logit. The profile is very close to the true-token profile.}
\label{fig:median_pred}
\end{figure}

The comparison between true-token and predicted-token logits is also stable.
Table~\ref{tab:true-vs-pred-fits} reports the fitted power law exponents and floors for
%\texttt{gradnorm\_diag} and \texttt{pred\_gradnorm\_diag}.
$\overline G^{\mathrm{true}}(r)$ and $\overline G^{\mathrm{pred}}(r)$.
The profiles are
nearly identical; see also Figure~\ref{fig:median_pred}. Thus the decay is not an artifact of differentiating the
corpus token even when the model would have predicted another token.

\begin{table}[htbp]
\centering
\resizebox{.7\textwidth}{!}{%
\begin{tabular}{lrrrr}
\toprule
Experiment
& $p_{\mathrm{true}}$
& $\gamma_{\mathrm{true}}$
& $p_{\mathrm{pred}}$
& $\gamma_{\mathrm{pred}}$ \\
\midrule
Pythia-14M, Gutenberg  & 0.8825 & --       & 0.8816 & -- \\
Pythia-70M, Gutenberg  & 0.8817 & 0.000971 & 0.8843 & 0.000996 \\
Pythia-160M, Gutenberg & 0.8193 & 0.004573 & 0.8212 & 0.004653 \\
Pythia-410M, Gutenberg & 0.7148 & 0.010076 & 0.7134 & 0.010744 \\
Qwen2.5-0.5B, Gutenberg & 0.8811 & 0.001787 & 0.8869 & 0.001809 \\
Pythia-14M, WikiText-103 & 0.8725 & -- & 0.8728 & -- \\
% Pythia-14M, C++ code & 0.9105 & -- & 0.9104 & -- \\
Pythia-14M, shuffled Gutenberg & 0.8372 & -- & 0.8347 & -- \\
\bottomrule
\end{tabular}%
}
\caption{Comparison between true-token and predicted-token sensitivity fits.
The table reports the %AIC-
logRMSE-
selected exponent and floor for
%\texttt{gradnorm\_diag} and \texttt{pred\_gradnorm\_diag}.}
    $\overline G^{\mathrm{true}}(r)$ and $\overline G^{\mathrm{pred}}(r)$.}
\label{tab:true-vs-pred-fits}
\end{table}

\subsection{Token shuffling}

The shuffled Gutenberg experiment is the most important
argument
%warning
against the
interpretation ``Green's function of text.''  Shuffling destroys coherent word order and
therefore destroys meaningful next-token prediction.
Nevertheless, over the measured distance range, the median Jacobian profile remains power-law-like, with a fitted exponent
%Nevertheless, the median
%Jacobian profile remains algebraic, with
\[
  p_{\mathrm{shuffled}}\approx 0.84.
\]
The amplitude and individual profiles are not identical to the coherent-text
case, but the qualitative long-range decay survives.

This means that the measured profile cannot be explained only by linguistic
coherence, syntax, or long-range semantic dependencies in the corpus.  The text
still matters: it supplies the token identities, embeddings, positions, and the
point at which the nonlinear network is linearized.

But the persistence under
shuffling shows that the power law profile is largely a property of the trained
operator acting on token sequences and
not determined solely by well-formed text.

%not %simply
%a property of well-formed text.

\subsection{Randomly initialized models}

The random-initialization test case %control
addresses the opposite possibility: perhaps
any Transformer with causal attention and positional structure would produce a
similar profile.  This is not observed.  A randomly initialized Pythia-14M model
with the same architecture has essentially no useful prediction ability, and its % off-diagonal
normalized sensitivity profile $\overline G^{\mathrm{true}}(r)$ is flat and extremely small.  The fitted power
exponent collapses to
$
  p\approx 0,
$
and the typical off-diagonal level is approximately
$
  \overline G^{\mathrm{true}}(r)\approx 7.6\cdot 10^{-4}.
$
The same conclusion holds for coherent Gutenberg input and shuffled Gutenberg
input.

Thus the power law decay is not a trivial consequence of the untrained
architecture.  Training is essential.  The learned weights create a long-range
Jacobian structure that is absent, or at least not visible at the same scale, in
the randomly initialized network.

\section{Interpretation for Coarse Spaces in Language Models}

%The operator-learning motivation
%arising from our recent work~\cite{Koehler:2026:HAD}
%should %can
%now be stated more precisely.
%In a PDE setting, a Green kernel tells us how perturbations propagate through the inverse operator.
%If this propagation is essentially local, then local blocks, sparse approximations, or one-level methods may be sufficient.
%If the response has a slowly decaying tail, then a coarse level is needed to represent the weak global interactions.
%%This is the standard logic behind two-level Schwarz methods, multigrid, and related hierarchical preconditioners.
%This is the standard logic behind two-level Schwarz methods, multigrid, and related hierarchical preconditioners~\cite{Lions1988,ToselliWidlund2005,SmithBjorstadGropp1996,trottbook}.
%Recent work at the interface of machine learning and domain decomposition further supports the idea that hierarchical numerical concepts can inform learned architectures~\cite{Klawonn2024,Koehler:2026:HAD}.

The operator-learning motivation arising from our recent work~\cite{Koehler:2026:HAD} can now be stated more precisely.
In a PDE setting, a Green kernel describes how perturbations propagate through the inverse operator.
If this propagation is essentially local, then independent local blocks, sparse approximations, or one-level methods may be sufficient.
If the response has a slowly decaying tail, then the long-range part should not be ignored -- but it also need not be represented by a dense global operator.
The typically more efficient hierarchical alternative is to combine local operators with a coarse representation of the weak global interactions.
This is the standard logic behind two-level Schwarz methods, multigrid, and related hierarchical preconditioners~\cite{Lions1988,ToselliWidlund2005,SmithBjorstadGropp1996,trottbook}.
%In our attention setting, this corresponds to replacing fully global attention by local attentions plus a coarse attention mechanism, e.g., as in~\cite{Koehler:2026:HAD}, where better approximation quality was achieved using fewer parameters by an order of magnitude in the hierarchical model compared to the standard attention.
%This report attempts to give a justification for this approach.

In the attention setting, the analogous idea is to replace fully global attention by a hierarchical approximation consisting of local attentions and a coarse attention mechanism. In~\cite{Koehler:2026:HAD}, this led to better approximation quality with about an order of magnitude fewer parameters than the corresponding standard global attention model. The present report provides empirical motivation for this hierarchy by measuring how input-output sensitivity in trained autoregressive language models decays with causal distance.

%Our measurements suggest an analogous, but not identical, situation for trained
%LLMs.  The model has strong local token interactions, but it also has weak
%global interactions that decay slowly with causal distance.  These interactions
%are not simply attention weights and not simply semantic relations between words.
%They are entries, or rather block norms, of the local Jacobian of the trained
%Transformer operator.

The model has strong local token interactions, but it also retains weak long-range interactions whose median strength decays only slowly with causal distance.  These interactions are not attention weights and not direct semantic relations between words.  Rather, they are sensitivities of a selected next-token logit with respect to earlier input-token embeddings.  Since each token is represented by a high-dimensional embedding vector, the derivative with respect to one earlier token is not a single scalar entry, but a small Jacobian block.  In our distance-resolved profile, this block is reduced to a scalar by taking its Euclidean norm.

This observation may help % how one might
to design
%think about
coarse variables for NLP.
A purely linguistic coarse space might be based on sentences, paragraphs,
topics, or summary tokens.  Such choices may be useful, but they are not the
only possibility.  From the Green-function view, a coarse space should represent
slowly decaying response components of the trained operator.  These modes may combine
position, token identity, syntax, and model-internal learned structure.  They
may therefore be only partially aligned with human semantic units.

This is also why the shuffling result is important,
%although it disappointing seemed
rather than disappointing.
If the profile survived only on coherent text, one might conclude that the
coarse level should be primarily semantic.  Since the power law profile survives
shuffling but disappears for random weights, the long-range component appears to
be a learned property of the trained sequence operator itself.  A future
hierarchical architecture inspired by
numerical analysis and scientific computing could %should therefore try
attempt
to capture this learned long-range Jacobian structure, not merely summarize the
surface text.

\section{Conclusion}

We asked how token influence decays with distance in trained language models,
motivated by Green functions, coarse spaces, and hierarchical preconditioners in numerical analysis.
%operator learning.
The measured object is a distance-resolved Jacobian
sensitivity: the derivative of a selected next-token logit with respect to an
earlier input embedding.  This is not a Green's function of natural language in
the classical sense.  It is a Green-function view of the trained Transformer as
a nonlinear operator, evaluated at given token sequences.

Across trained Pythia models, Qwen2.5-0.5B, %coherent prose,
%WikiText-103%, and C++ code,
using text from Gutenberg and Wikitext-103,
the median diagonal-normalized sensitivity profile
%decays approximately according to a power law.
is well described, over the measured distance range, by a power-law-type fit.

The profile persists under token shuffling, although token
prediction quality collapses, but disappears for randomly initialized models.
Thus the main phenomenon appears to be a learned long-range response of trained
Transformer LLMs.

Existing hierarchical, sparse, retrieval-based, or token-compression mechanisms in the LLM literature---including summary tokens, pooling tokens, memory tokens, and related global-token constructions---are typically motivated by efficiency, architectural, or semantic considerations.

%Existing hierarchical, sparse, or retrieval-based mechanisms in the literature of LLMs are typically motivated by efficiency, architectural, or semantic considerations.
In contrast, we motivate hierarchy from the measured, distance-resolved Jacobian
structure of the trained operator.
For hierarchical model design, this suggests that in NLP coarse spaces can be designed using similar principles of locality and global coupling as in numerical analysis and scientific computing, as in our own attempt in~\cite{Koehler:2026:HAD}, which motivated this work.

\begin{remark}
It was our original goal to identify a Green's function for text.  However, our
experiments show that this is not quite what we have obtained.  Token shuffling
destroys coherent word order and makes next-token prediction much worse, but
the distance-resolved token influence profile remains similar to the case of
standard text.  We also find that randomly initialized Transformer models do
not show the same distance profile.  Thus, what we measure %the measured object
is not directly
an intrinsic property of natural language text, and it is also not a trivial
property of the Transformer architecture.  Rather, it appears to be a property of the
trained Transformer operator, acquired during training on natural language.  In
this sense, the experiment views natural language through the lens of the
trained model: the token sequence specifies where the nonlinear operator is
linearized, while the learned weights determine the response.
\end{remark}

%that the analog of a coarse space in NLP
%should not be chosen solely from text structure.  It should also account
%for slowly decaying Jacobian response %modes (todo: not modes, Jacobian response)
%of the trained operator.
%
%In this sense, the
%Green-function question remains central: it tells us what kind of global
%coupling a coarse level would need to represent.

\section*{Statement on AI Use}
The large language models ChatGPT 5.5, Grok 4.5 and Mistral Large 3 675B Instruct 2512 were used to assist in the literature review, the development of the Python codes for the methods discussed in this paper and for the data extraction from the log files.
Mistral was used via the Chat-AI service of the GWDG~\cite{doosthosseini_saia_2026}.
%The authors reviewed, tested, debugged, and iteratively refined the generated code.
%LLMs were also used for the data extraction from the log files.
%The numerical experiments reported here were performed with one of these prototypes and checked against the independently developed second prototype.
The LLMs were also used to support the preparation of the manuscript, in particular to generate the tables, improve wording, structure, and presentation.

\bibliographystyle{unsrtnat}
\bibliography{./airefs}

\end{document}